\documentclass[runningheads]{llncs}
\usepackage[T1]{fontenc}
\usepackage{graphicx}
\usepackage[usenames, dvipsnames, svgnames, table]{xcolor} 

\usepackage[framemethod=tikz]{mdframed} 

\newcommand{\B}{\bfseries}

\usepackage{amsmath}
\begin{document}
\title{Image Translation Based Nuclei Segmentation for Immunohistochemistry Images}
\titlerunning{Nuclei Segmentation for Immunohistochemistry Images}
%
\author{Roger Trullo\inst{1} \and
Quoc-Anh Bui\inst{2} \and
Qi Tang\inst{3} \and
Reza Olfati-Saber\inst{4}}
\authorrunning{R. Trullo et al.}



%
\institute{Sanofi, Chilly Mazarin, France \and
Aix-Marseille University, Marseille, France \and
Sanofi, Bridgewater, NJ, USA \and
Sanofi, Cambridge, MA, USA\\
\email{roger.trullo@sanofi.com}\\
\email{quoc-anh.bui@sanofi.com}\\
\email{qi.tang@sanofi.com}\\
\email{reza.olfati-saber@sanofi.com}}


%
\maketitle              
\begin{abstract}
Numerous deep learning based methods have been developed for nuclei segmentation for H\&E images and have achieved close to human performance. However, direct application of such methods to another modality of images, such as Immunohistochemistry (IHC) images, may not achieve satisfactory performance. Thus, we developed a Generative  Adversarial Network (GAN) based approach to translate an IHC image to an H\&E image while preserving nuclei location and morphology and then apply pre-trained nuclei segmentation models to the virtual H\&E image. We demonstrated that the proposed methods work better than several baseline methods including direct application of state of the art nuclei segmentation methods such as Cellpose and HoVer-Net, trained on H\&E  and a generative method, DeepLIIF, using two public IHC image datasets.

\end{abstract}
\section{Introduction}
\begin{figure}
\includegraphics[width=\textwidth]{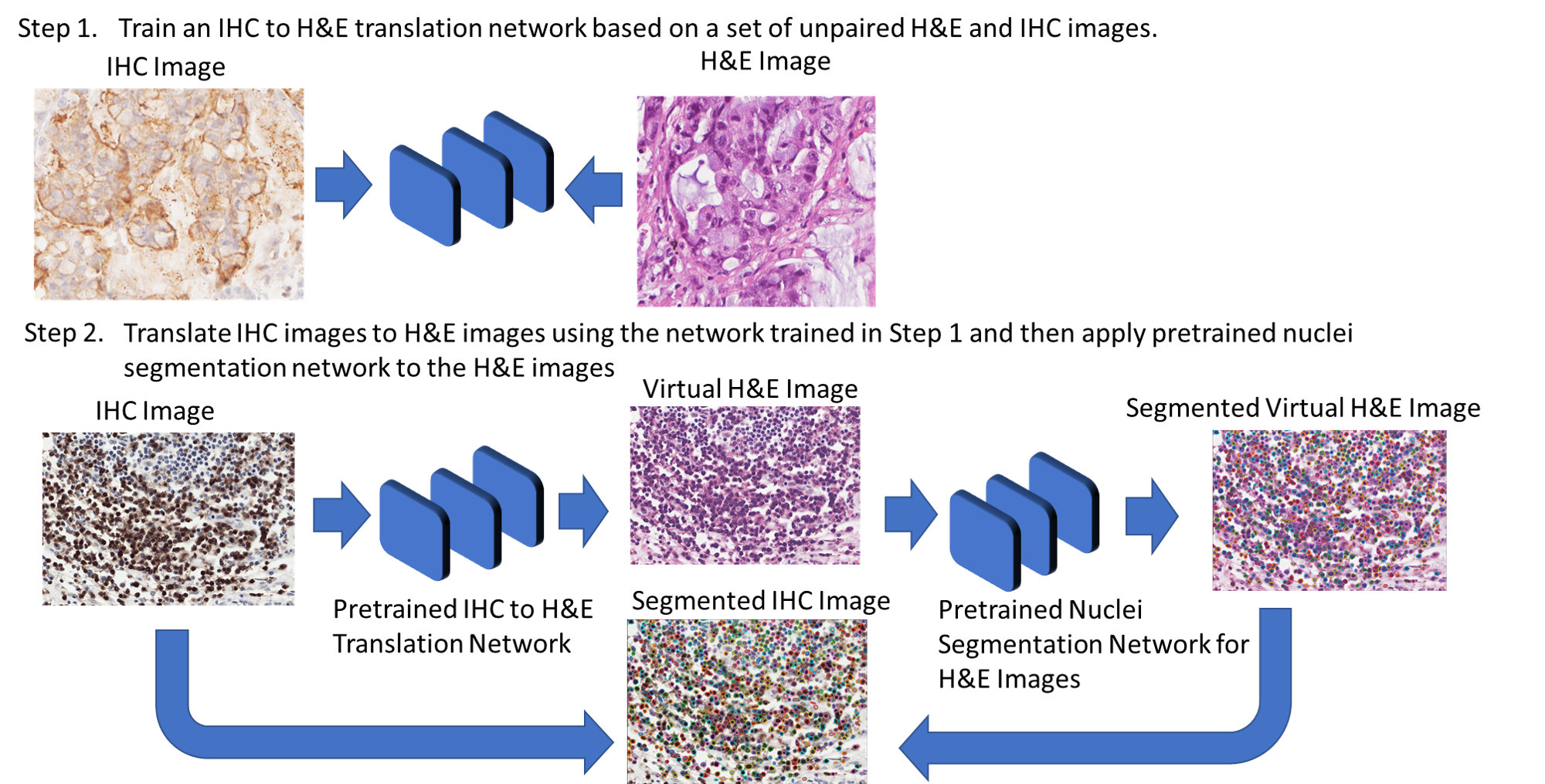}
\caption{Proposed IHC Segmentation Pipeline.} \label{pipeline}
\end{figure}

H\&E images are the most common modality of histopathology images since they can be stained quickly, economically and significance amount of microscopic anatomy is revealed. As a result, many nuclei segmentation methods have been developed recently leveraging pathologist's annotations and the advance in computer vision for H\&E images~\cite{mahbod2021cryonuseg}. Besides H\&E, there is another popular type of image, the immunohistochemistry (IHC) image, which is commonly used to identify specific protein biomarkers and is complementary to H\&E images~\cite{kaplan2015quantifying}.  IHC images play  a central role in companion diagnostic tools for development of precision medicines. Several novel oncology therapies have been approved by regulatory agencies with a companion diagnostic device based on IHC images. On the other hand, the analysis of IHC images has been typically based on manual semi-quantitative methods, such as the 20X rule, for clinical decision making, which may suffer from subjectivity and inter and intra-rater variability. Several case studies suggest that deep learning based digital pathology algorithms may provide improved patient selection comparing to the current clinical standard~\cite{glass2021821}. At the core of these deep learning algorithms, there is a nuclei segmentation method, which relies on human expert annotations. 
To avoid the labor intensive manual annotation on IHC images and also to improve upon the performance of direct application of models trained on H\&E images to IHC images, we proposed a  two-step label-free  approach where an H\&E image is first translated into an IHC image utilizing unsupervised image-to-image translation methods, then for the second step, we apply existing methods that performed well on H\&E images to  virtually generated H\&E images to obtain nuclei segmentation masks (Fig. \ref{pipeline}). 

\section{Related Work}
Numerous  cell morphology operation methods, such as color deconvolution,  Otsu's thresholding, watershed, have been used either separately or in combination to segment nuclei in histology images~\cite{di2010automated}. With the advent of deep learning, it has been shown that supervised methods can outperform traditional cell morphology based methods~\cite{yang2020nuset} and traditional machine learning based methods~\cite{hatipoglu2017cell}. However, a drawback of these methods and the supervised learning methods in general is that labor intensive manual annotation is required by trained human experts~\cite{abdolhoseini2019segmentation}. To avoid manual annotations, unsupervised deep learning methods have been developed~\cite{han2021unsupervised,chen2019unsupervised}  for nuclei segmentation, however these general methods were not tailored for IHC images.  To develop nuclei segmentation models for IHC images with few annotations, researchers have leveraged the power of multiplex immunofluorescence (IF) staining to provide a ground truth of nuclei masks for the IHC nuclei segmentation through co-registration of IF images with IHC images~\cite{ghahremani2022deepliif} and then trained a conditional generative adversarial network (cGAN)~\cite{isola2017image} to generate nuclei masks.  

\section{Methodology} 
\subsection{Translation from IHC Images to H\&E Images}
Image-to-image translation have achieved significant advancements since the advent of GAN. Application of GAN and its variations created a new field in digital pathology named virtual staining, where one or more image modalities can be virtually generated based on input of only one modality of image. Feasibility of virtual staining has been demonstrated across multiple histology image modalities~\cite{bayramoglu2017towards,rivenson2019virtual,xu2019gan,zhang2020digital,rivenson2020emerging} including H\&E and IHC. 
When it comes to translation between H\&E and IHC, almost all literature focused on the translation from H\&E to IHC since H\&E images were more commonly available for patients. However, for our purpose of leveraging well trained nuclei segmentation models for H\&E images, we are interested in the reserve direction of the translation, that is, from IHC to H\&E. 
Due to the challenges of registering IHC images with H\&E images and high resolution of these types of images, generative methods that do not require paired images, such as CycleGAN~\cite{zhu2017unpaired} and U-GAT-IT~\cite{Kim2020U-GAT-IT}, are methods of interest for the image translation task. These methods work by training simultaneously two GANs (two generator and two discriminator models). One generator translates IHC to H\&E and other generator translates from H\&E to IHC. In addition, in order to have pixel to pixel correspondence, a cycle consistency loss is introduced. This loss enforces the idea that when using as input an IHC image, the first generator will produce a virtual H\&E image and if we use this virtual H\&E image as input to the second generator, the produced image should match the input IHC image. On the other hand, U-GAT-IT extended the method by adding new normalization layers and by using attention modules which was claimed to  improve the quality of the produced images.
For both of these methods,  we used the hyperparameters recommended by their authors including loss weights and learning rates. For CycleGAN we used a batch size of 10, whereas for U-GAT-IT we used a batch size of 2. This is due to the fact that U-GAT-IT uses more memory since it has fully connected layers. We trained CycleGAN for 30 epochs and U-GAT-IT for 20 epochs. The batch size was selected to use the full GPU memory available and the number of epochs was selected empirically by looking qualitatively at the generated virtual H\&E tiles. We used an NVIDIA P100 GPU for training of the image translation models, which took around 4 hours.


\subsection{Nuclei Segmentation on Virtual H\&E Images}
Once virtual H\&E images are generated, pretrained models that work well for H\&E images can be applied to the virtual H\&E images to obtain the nuclei segmentation masks. Since the image translation step kept the location and the morphology of each nuclei from the IHC images, the virtual H\&E images  perfectly matched with the IHC images at pixel level without the need of image registration.  Thus, the nuclei segmentation masks obtained from the virtual H\&E images can be directly used  as the nuclei masks for the original IHC images. Without loss of generality, popular H\&E image nuclei segmentation methods such as Cellpose~\cite{stringer2021cellpose}, StarDist~\cite{schmidt2018},  and HoVer-Net~\cite{graham2019hover}, were considered for this step. 
Cellpose was trained on a diverse set of datasets including image sets BBBC038v1 and BBBC039v1~\cite{caicedo2019nucleus},  image set MoNuSeg~\cite{kumar2017dataset}, and image set ISBI 2009~\cite{coelho2009nuclear}. The mixed dataset consists of microscopy images, H\&E images and fluorescence images, with about 1139 images for the task of nuclei segmentation. 
StarDist  was trained on two H\&E image datasets,  MoNuSeg 2018 training dataset, which has 30 H\&E images and around 22,000 nuclear boundary annotations from  a diverse set of patients including breast cancer, kidney cancer, lung cancer, prostate cancer, bladder cancer, colon cancer and stomach cancer patients,  and a TNBC dataset~\cite{naylor2018segmentation}, which has 50 annotated H\&E images from patients with triple negative breast cancer.  Regarding HoVer-Net,  it was trained on 41 H\&E stained colorectal adenocarcinoma image tiles, containing 24,319 exhaustively annotated nuclei with associated class labels.
\section{Experiments}
We systematically evaluated both the image translation component and the nuclei segmentation component of the proposed approach to examine the impact of each component in comparison to two different types of baseline methods, the direct application of the nuclei segmentation models trained on H\&E images to IHC images and an pre-trained nuclei segmentation model tailored for IHC images, DeepLIIF~\cite{ghahremani2022deepliif}.  For all the experiments we used a P100 Nvidia GPU. 
\subsection{Datasets}
The image translation component  was trained based on an in-house IHC/H\&E dataset.  The nuclei segmentation component was tested and compared on two different IHC image datasets, the DeepLIIF testing dataset~\cite{ghahremani2022deepliif} and the LYON19 dataset~\cite{swiderska2019learning}.

\subsubsection{In-house IHC/H\&E Dataset}
To train the generative model that transforms IHC images onto virtual H\&E images we use an in-house dataset, that consists of 123 IHC whole slide images and 121 H\&E whole slide images. The IHC images were stained to highlight cells expressing a protein target, which cannot be disclosed due to confidentiality reasons, however, it is shown in Fig. \ref{fig:Visualize_cases} that the generative model trained on this specific type of IHC dataset generalized well to other protein targets, such as CD3/CD8 and Ki67. For each of these slides, we randomly sample patches of size $256 \times 256$ and ended up with  2510 patches from IHC images and 2793 patches from H\&E images.



\subsubsection{DeepLIIF Testing Dataset}\label{sec:deepliif_dataset}
A public test set \cite{ghahremani2022deepliif}, which can be downloaded from \url{https://zenodo.org/record/5553268}, it includes 598 Ki67 IHC images of size 512×512 and 40x magnification from bladder carcinoma and non-small cell lung carcinoma slides. The expression of Ki67 is strongly associated with tumor cell proliferation and growth, and is widely used in routine pathological investigation as a proliferation marker.

\subsubsection{LYON19 Dataset}\label{sec:LYON19}
Public testing set of LYON19 \cite{swiderska2019learning}, which can be downloaded from \url{https://zenodo.org/record/3385420}, it contains 441 Regions of Interest (ROIs) from whole slide images (WSIs) of CD3/CD8 stained IHC images of lymphocytes from breast, colon, and prostate cancer patients' biopsy specimens. The 441 ROIs were selected from IHC images with 277 regular area ROIs, 59 clustered cell ROIs, and 105 artifact area ROIs. The ground truth of this dataset was not disclosed by the Grand Challenge competition. However, performance on this testing set can be obtained after submitting the predictions to the challenge here \url{https://lyon19.grand-challenge.org/Submission/}. The objective of the competition is to detect positive stained cells; in particular, the center of each positive cell.
To accommodate for this, we computed our regular pipeline and then applied a simple thresholding procedure based on the HSI color space to classify cells as positively stained. We used the open source implementation provided by HistomicsTK \cite{histomicstk}. Finally, we computed the centroid of each of those cells.

\subsection{Baseline Methods and Evaluation Metrics}
We compared the proposed methods with four baseline models: the pre-trained Cellpose model v2.0.5~\cite{Cellpose_v2.0.5}, the pre-trained \texttt{2D\_versatile\_he} StarDist model v0.8.2~\cite{StarDist_0.8.2}, the pre-trained HoVer-Net model v1.0~\cite{hover_net_1.0}, and the pre-trained DeepLIIF model v1.1.2~\cite{DeepLIIF_1.1.2}. 

The performance of all the methods will be evaluated based on Dice score, which measures pixel level segmentation performance. We also evaluated cell instance level performance metrics including accuracy, precision, recall and F1 score, conditional on a given Intersection over Union (IoU) threshold. Since the concept of true negatives in instance cell level detection is not valid, accuracy is computed from the number of true positives, TP, false positives, FP and false negatives, FN, as accuracy$=\mbox{TP}/(\mbox{TP}+\mbox{FP}+\mbox{FN})$ as it is commonly done in object detection. 

\subsection{Results}
\begin{figure}
\includegraphics[width=\textwidth]{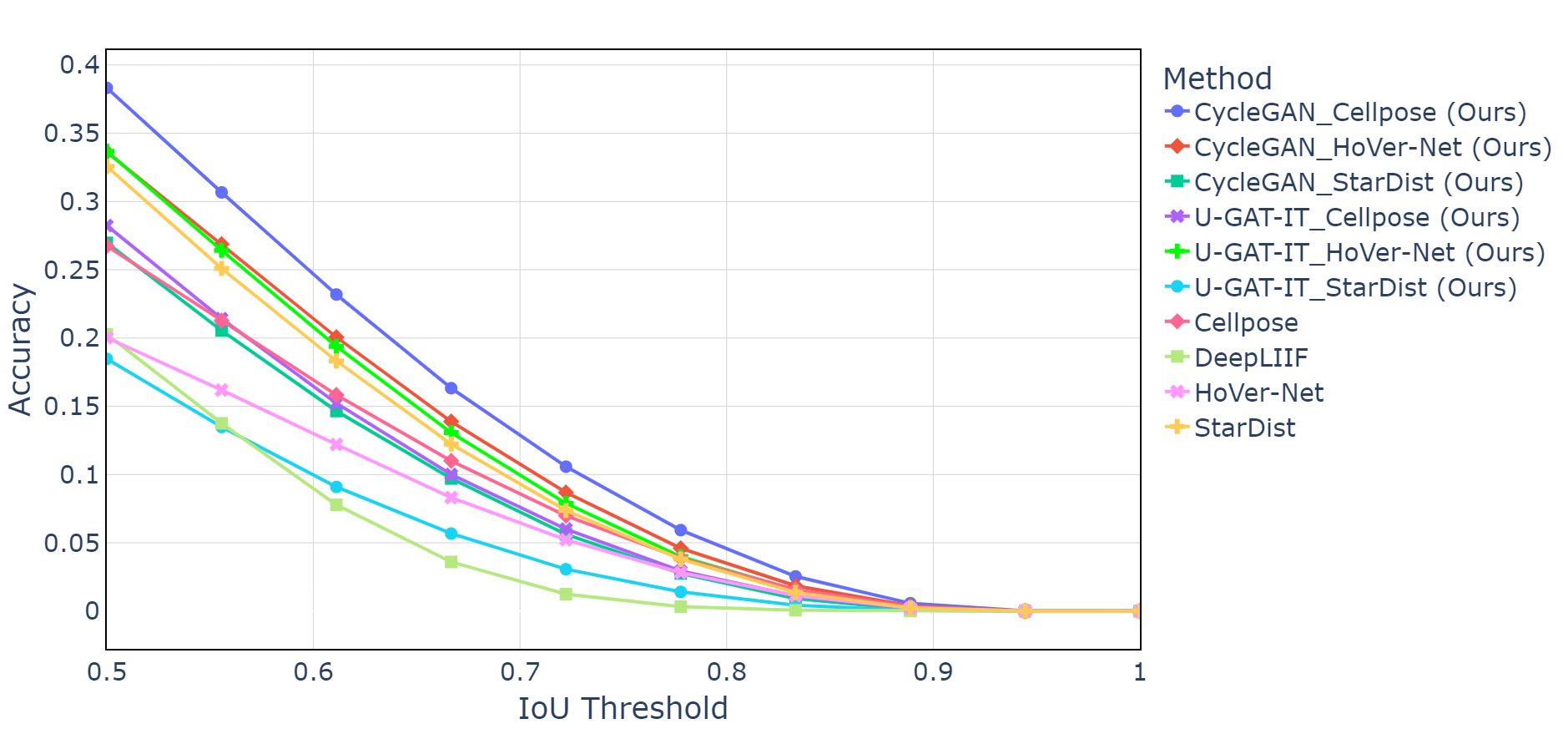}
\caption{Segmentation accuracy curves for proposed methods and baseline methods when IoU changes from 0.5 to 1 with a step size of 0.05.} \label{fig:DeepLIIF}
\end{figure}

First, the proposed methods were compared against the baseline methods on the DeepLIIF testing dataset. CycleGAN plus Cellpose achieved the best results in terms of Dice score and precision at IoU$=0.5$ as shown in Table \ref{tab:DeepLIFF_results}.  By translating IHC to H\&E images using GAN based methods, such as CycleGAN and U-GAT-IT, the performance of Cellpose and HoVer-Net improved at least 0.1 in Dice score comparing to directly application of these methods to IHC images. However, the combination of GAN based methods with StarDist failed to achieve improvement and even lead to worse performance comparing to StarDist alone. We also observed that CycleGAN performed better than or similarly to U-GAT-IT when combined with StarDist, Cellpose or HoVer-Net. By changing the thresholds for IoU, the accuracy curves in Fig. \ref{fig:DeepLIIF} confirm that the same conclusion holds against different thresholds.

\begin{table}
\caption{Performance of proposed methods against baseline methods on DeepLIIF testing dataset. Cell instance level segmentation accuracy, precision, recall and F1 score were evaluated at $\mbox{IoU}=0.5$.}\label{tab:DeepLIFF_results}
\begin{tabular}{|r|r|r|r|r|r|}
\hline
Method      &  Dice Score & Accuracy & Precision & Recall & F1 Score\\
\hline
DeepLIIF            &  0.66 &   0.20 &  0.31    & 0.37  & 0.34 \\
StarDist            &  0.64 &   0.33 &  0.54    & 0.45  & 0.49  \\
CycleGAN+StarDist   & 0.59  &   0.27 &  0.49    & 0.38  &  0.42 \\
U-GAT-IT+StarDist   & 0.50 &    0.18  & 0.34    & 0.28  & 0.31  \\
Cellpose            &  0.57 &   0.27 &  0.58    & 0.33  & 0.42  \\
CycleGAN+Cellpose   &\B 0.72& \B 0.38& 0.63     &\B 0.49&\B 0.55  \\
U-GAT-IT+Cellpose   & 0.67  &   0.28 &   0.47   &  0.42  & 0.44  \\
HoVer-Net           & 0.42 &     0.20 &\B 0.64  &  0.23   & 0.33   \\
CycleGAN+HoVer-Net & 0.68 &     0.34  & 0.62    &  0.42   & 0.44 \\
U-GAT-IT+HoVer-Net & 0.68 &     0.34  & 0.59    &  0.44  & 0.50\\
\hline
\end{tabular}
\end{table}

\begin{figure}
 \includegraphics[width=1\textwidth]{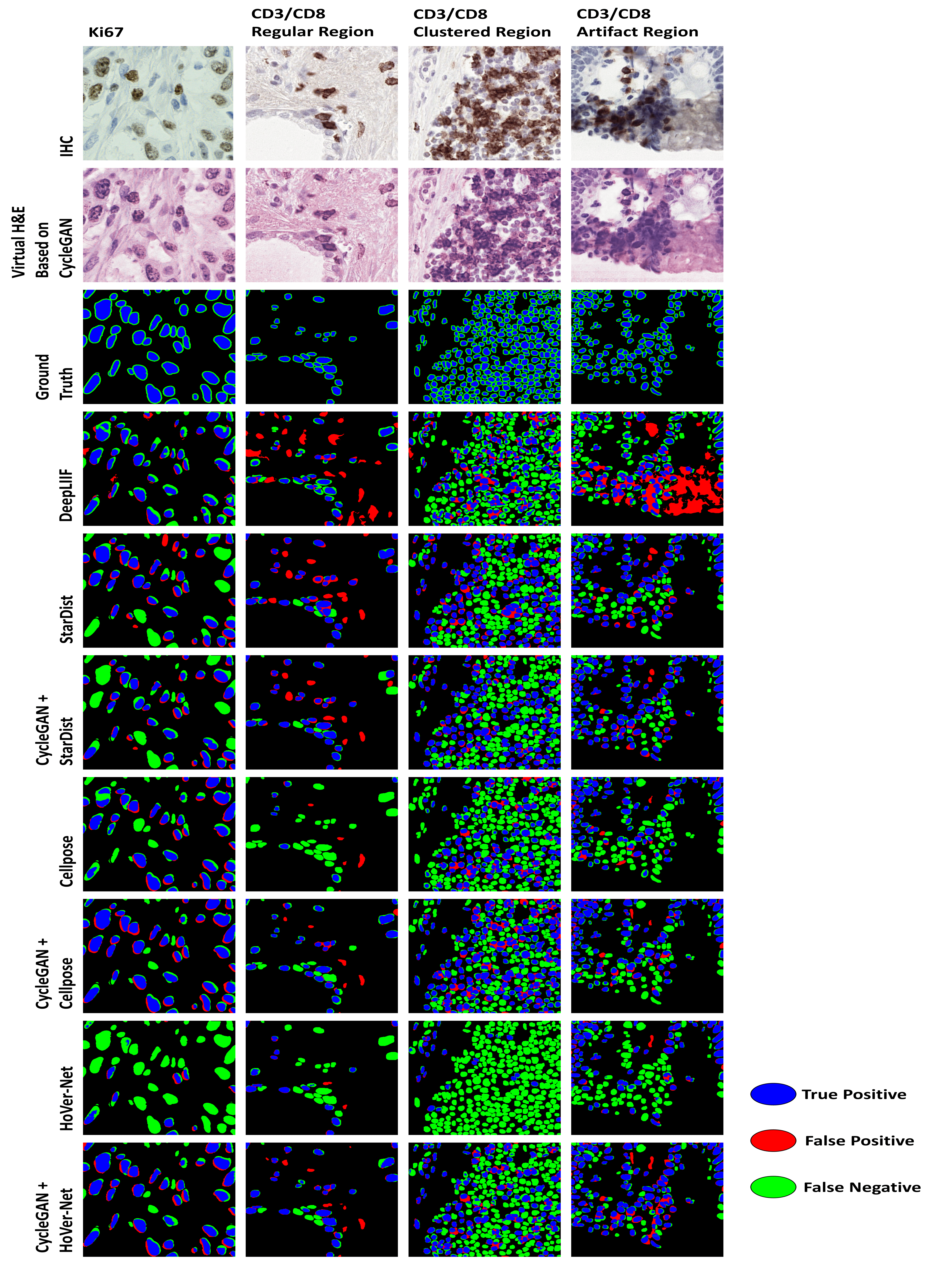}
\caption{Presentation of the proposed and the baseline methods' performance in Ki67 IHC images from DeepLIIF testing dataset and CD3/CD8 IHC images from LYON19 dataset.}
\label{fig:Visualize_cases}
\end{figure}

Fig. \ref{fig:Visualize_cases} visualizes cell nuclei segmentation results of four selected tiles. From the top to the bottom are the input IHC, the virtually generated H\&E based on CycleGAN, the ground truth, which was only available for the visualized tiles, and the cell masks of all the methods previously presented. Pixel-wise, the true positive (TP), false positive (FP) and false negative (FN) are represented by blue, red and green color respectively.
The first row presents a testing Ki67 IHC image from the DeepLIIF testing set whereas the next 3 are the CD3/CD8 IHC images from the LYON19 dataset. For the Ki67 IHC image, the proposed methods perform similarly to DeepLIIF since DeepLIIF was trained specifically for this type of staining. However, for the CD3/CD8 IHC images, the proposed methods outperformed DeepLIIF and the direct application of Cellpose and HoVer-Net models pretrained on H\&E images.

Next, we examined the performance of these methods on the LYON19 dataset. By submitting the model predictions to the Grand Challenge competition website, the F1 score were obtained and were tabulated in Table \ref{tab:LYON19_results}. Only CycleGAN was used for the image translation step since it has better performance than U-GAT-IT when tested in the DeepLIIF testing dataset. The proposed method performed similarly as the baseline methods except the combination of CycleGAN with HoVer-Net method, which lead to 4\% higher F1 score in all detections, 9\% higher F1 score in clusterd cells and 8\% higher F1 score in artifact areas comparing to the DeepLIIF method. When compared against the HoVer-Net itself, the improvement in F1 score is much larger with at least 20\% higher F1 scores across all categories. 

\begin{table}
\caption{F1 scores for different types of segmentation tasks of proposed methods against baseline methods on LYON19 dataset.}\label{tab:LYON19_results}
\begin{tabular}{|r|r|r|r|r|}
\hline
Method      &  All Detections & Regular Areas & Clustered Cells & Artifact Areas\\
\hline
DeepLIIF            &  0.53 & \B  0.64 &  0.59    & 0.17  \\
StarDist            &  0.54 &   0.59 &  0.63    & 0.22  \\
CycleGAN+StarDist   & 0.54  &   0.59 &  0.63    & 0.22  \\
Cellpose            &  0.54 &   0.53 & \B 0.68    & 0.24  \\
CycleGAN+Cellpose   &  0.54 &   0.61 &  0.67    & 0.20   \\
HoVer-Net           & 0.04 &     0.06 & 0.01  &  0.03     \\
CycleGAN+HoVer-Net &\B 0.58 &     0.63  &\B 0.68    & \B 0.25  \\
\hline
\end{tabular}
\end{table}

\section{Conclusions}
We leveraged publicly available pre-trained nuclei segmentation models based on H\&E images to perform nuclei segmentation in IHC images without requiring any manual anotation on them. This was achieved by virtually translating IHC images into H\&E images. To enable such translation, we trained image translation models, such as CycleGAN and U-GAT-IT models, based on an in-house dataset of IHC and H\&E images. This approach achieved better performance than several of the baseline methods, including direct application of pre-trained models based on H\&E images, such as Cellpose and HoVer-Net, to IHC images,  and a pre-trained model tailored for IHC image nuclei segmentation,  DeepLIIF, when tested on DeepLIIF testing dataset. Such improvement was less pronounced in the LYON19 dataset except for CycleGAN combined with HoVer-Net. An interesting finding is that HoVer-Net alone performed badly on LYON19 dataset however when combined with CycleGAN, it became the best performing method. These observations together with the findings that the combination of CycleGAN or U-GAT-IT with the StarDist method failed to achieve improved performance in both testing datasets suggest that if the pre-trained nuclei segmentation method for H\&E images has strong generalizability to IHC images, translating IHC to H\&E images may not lead to improved performance, such as the case for StarDist, however, on the other hand,  if the pre-trained method utilized features specific to H\&E images, image translation can lead to substantial improvement. Thus, adoption of the proposed method depends on the generalizability of the pre-trained models to the target image modality. 
Our work has been based completely on label-free IHC images. An interesting perspective is to combine our method with a few manual annotations in a semi-supervised learning setting which we believe can potentially improve the performance even further.
%
%
%
\bibliographystyle{splncs04}
\bibliography{bibliography.bib}
%




\end{document}